# An Accurate Arabic Root-Based Lemmatizer for Information Retrieval Purposes

Tarek El-Shishtawy[1] and Fatma El-Ghannam[2]

[1] Benha University
Cairo, Faculty of Engineering (Shoubra) , Egypt
*shishtwy@hotmil.com*

[2] Electronics Research Institute
Cairo, Egypt
*elghannamf@hotmail.com*

**Abstract**

In spite of its robust syntax, semantic cohesion, and less ambiguity, lemma level analysis and generation does not yet focused in Arabic NLP literatures. In the current research, we propose the first non-statistical accurate Arabic lemmatizer algorithm that is suitable for information retrieval (IR) systems. The proposed lemmatizer makes use of different Arabic language knowledge resources to generate accurate lemma form and its relevant features that support IR purposes. As a POS tagger, the experimental results show that, the proposed algorithm achieves a maximum accuracy of 94.8%. For first seen documents, an accuracy of 89.15% is achieved, compared to 76.7% of up to date Stanford accurate Arabic model, for the same, dataset.

## 1. Introduction

In the field of NLP, lemmatization refers to the process of relating a given textual item to the actual lexical or grammatical morpheme [9]. Both of the word representation granularity level and its extracted morpho-syntactic features directly affect the performance of Information Retrieval (IR), Machine Translation, Summarization and keyphrase extraction systems. In semitic languages, such as Arabic, this is not an easy task due to their highly inflectional properties.

The granularity level is determined by clustering different words into groups, which shares the same root, stem or lemma. While many IR researchers consider the root level as the basic group others raise the importance of stem level for improving the effectiveness of IR systems. Lemma level analysis and generation does not get yet much focus in spite of its robust syntax, semantic cohesion, and less ambiguity

Lemma refers to the set of all word forms that have the same meaning, and hence capture semantic similarities between words. Recent researches in Arabic IR systems show the need of representing Arabic words at lemma level for many applications, including statistical machine translation [10,13], keyphrase extraction [12], indexing and classification [17]. Lemmatization is relatively new topic for Arabic language processing, and hence only few researches focused directly on automatic lemma extraction from Arabic texts. Light stemmers and supervised learning approaches are the two main approaches for POS tagging and lemma extraction.

In spite of their limited accuracy, light stemmers and light lemmatizers introduce many useful techniques for disambiguating word category with minimum resources, which make them attractive to IR purposes. However, light stemmers fail in many cases to group related words [23], since there are no roots or stems to verify with. For example, it fails to conflate forms such as broken (irregular) plurals for nouns and adjectives with their singular forms, and past tense verbs do not get conflated with their present tense forms, because they retain some affixes and internal differences.

On the other hand, statistical supervised learning approaches present the best published accuracy as POS taggers. The cost of expanding language coverage is a major problem in supervised learning approaches. In closed learning methodologies, it is not an easy task to add new entries, since the whole model has to be retrained for these new entries. It is noted that [31], tagging accuracy for statistical approaches markedly decreases for previously unseen words. In our experimental results, the accuracy of -up to date- Stanford learning model for Arabic language was found to drop from its 96.86% best published result to only 76.7% for documents in education domain.

In this paper, we present a different approach that depends on Arabic language knowledge to disambiguate word category. We are motivated by the notion that the tagger performance can be improved by expanding the knowledge sources available to the tagger. In fact, Arabic language specialists are still able to disambiguate unknown words (e.g., newly added Arabized words) by different language knowledge resources. In this paper, the proposed lemmatizer uses word patterns, roots, syntactic and morphological basic rules, to reduce Arabic words into their lemma canonical form. In the proposed approach, the extraction process is augmented with auxiliary dictionaries for words that are expected to fail in analysis with rules. The proposed approach extracts also the lexical category (POS) of the input word. To test the effectiveness of the proposed approach, we compared the results by up to date Stanford Arabic accurate tagger model. The results show that in all cases, better accuracy is obtained.

The rest of this paper is organized as follows: section (2) gives a brief description of related works in Arabic stemming techniques. The basic features of the algorithm are presented in section 3. The proposed algorithm has two main phases; POS tagging phase and lemma generation phase. The two phases are reviewed in sections 4 and 5. We present in section 4 the methodology and tools for building the proposed light lemmatizer. Section 6 gives performance analysis of the proposed algorithm. Finally, section 7 concluded the results.

## 2. Related works

Arabic is very rich in categorizing words, and hence, numerous stemming techniques have been developed for morphological analysis and POS tagging. Stemming and lemmatization shares a common purpose of reducing words to an acceptable abstract form, suitable for NLP applications. Previous works have presented important approaches that reflect different points of views to the problem. This includes debated word representation level, direct versus rule based lexicons, light stemmers versus accurate supervised learning approaches. The following subsections review these main trends and their relation to IR purposes.

2.1 Word representation level

There is no general agreement of the representation level of Arabic words in IR systems. Two levels have been debated; root level and stem level. Motivated by the power of Arabic roots, the first approach represents words in their root forms. Historically, a root was the entry to traditional Arabic lexicons. Almost, every word in Arabic is originated from a root. Many researchers [6,8], have emphasized that the use of Arabic roots as indexing terms substantially improves the information retrieval effectiveness over the use of stems.

However, several researchers [1,10] criticized this approach, and based their representation on stems. A stem is the least marked form of a word that is the uninflected word without suffixes, prefixes. Stem-based algorithms, remove prefixes and suffixes from Arabic word until it matches one of the Arabic patterns, or generate such patterns from a root. Dichy and Fargaly [10], suggested stem level to be the basis of lexical entries of morpho-syntactic and semantic information. Also, Attia [5], concluded that using the stem as base form is far less complex in developing and maintaining, less ambiguous, and more suitable for syntactic parsers that aim at translation.

The main problem in selecting a root as a standard representation level in information retrieval systems is the over-semantic classification. Many words that do not have similar semantic interpretations are grouped into same root. For example, Arabic words ("مكتبة" library M123H), ("كاتب" writer 1A23) and ("كتاب" book 12A3) originate from the same root ("كتب" 123), while having different semantic senses. Thus, using the root-based algorithms in information retrieval may increase the word ambiguities.

On the other hand, stem level suffers from under-semantic classification. Stem pattern may exclude many similar words sharing the same semantic properties. For example, Arabic broken plurals have stem patterns which differs from their singular patterns. Stem-based representation, cannot detect the syntactic similarity between (طائر bird) and (طيور birds), since they have different stem patterns (1A23 and 12O3).

2.2 Direct and Rule-Based Lexicons

Direct access lexicons produce fewer errors, as they store a complete set of possible words, with their morpho-syntactic features. Early work on Arabic stemming used manually constructed dictionaries, till now this approach is still widely used. Al-Kharashi and Evens [2], worked with small text collections, for which they manually built dictionaries of roots and stems for each word to be analyzed. Buckwalter developed an Arabic Morphological Analyzer [7], as a set of lexicons of Arabic stems, prefixes, and suffixes, with truth tables indicating legal combinations. The main problem of direct access dictionary based approach is the dictionary size and the cost of maintenance.

In contrast, to direct accessed lexicons, other types of analyzers reduce the size of the corpus or dictionary by getting benefits from the derivational and inflectional productiveness of Arabic. Using morphological rules nouns and verb stems are derived from a few thousand roots by infixing. Such systems attempt to find the root, or any number of possible roots for each word. Xerox Arabic Morphological Analysis and Generation [6], is one of the superior rule based - root based systems. The system includes 4,930 roots and 400 patterns, effectively generating 90,000 stems.

Another superior open source root-based stemmer is the Khoja's stemmer [21]. A comparative study for three Arabic morphological analyzers and stemmers (Khoja stemmer, Buckwalter analyzer, and Tri-literal Root Extraction Algorithm), shows that Khoja stemmer achieves the highest accuracy [29].

The main criticisms to rule based approach are its coverage limitations, and abundance of many forms to reduce the ambiguity. For example, some of the most closely related forms such as singular and plural nouns are irregular, and are not related by simple affixing. For information retrieval, this abundance of forms means a greater likelihood of mismatch between the form of a word in a query and the forms found in documents relevant to the query. Another problem with rule based approach is that the generative power of rules makes it possible to produce forms that are ambiguous or unknown in the language.

## 2.3 Light Stemmer Approaches

Many statistical based approaches of IR systems rely on surface form of the word. This was the motivation for light stemming approaches. Light stemmers, are not an aggressive practice as the root-based algorithms. The aim of the light stemmers is not to produce the linguistic root of a given Arabic surface form, rather it is to remove the most frequent suffixes and prefixes [18]. Larkey's light stemmer [23], removes blindly the most frequent suffixes and prefixes from Arabic surface word given to produce the stem.

Hammouda et.al. [17], presented an approach to generate Arabic lemma for indexing purposes. The algorithm was based on removing suffixes, prefixes and all vowels from Arabic words without checking the validity of generated lemma. A similar heuristic approach was presented by Al-Shammari et.al. [3]. Their algorithm searches text words, hoping to find one of predetermined specific stop words, to differentiate between verbs and nouns and hence selects the appropriate stemming process. Both of the two algorithms can be categorized as light lemmatizers,

In spite of their limited accuracy, light stemmers and limmateziers introduce many useful techniques for disambiguating word category with minimum resources, which makes them attractive to IR purposes. However, light stemmers does not guarantee that the extracted lemma is a real word in Arabic language and suffers from limited extracted features, Moreover, light stemmers fail in many cases to group related words, since there are no roots or stems to verify with. For example broken plurals do not get conflated with their singular forms, and past tense verbs do not get conflated with their present tense forms, because they have internal differences in word structure**.**

## 2.4 Supervised Learning Approaches

Accurate analyzers cannot rely only on dictionary to disambiguate word categories. Many stems with different POS tags can be produced as possible stems of a given word. This problem exists in most languages; however it is serious in Arabic. For example, the word (عَلَم teach) and (علم science) and (علم flag) are entries for same written Arabic word, as Arabic words are not normally written in their diacritic forms.

Many researchers deal with lemma extraction as a supervised learning problem. In general, these approaches give the best known accuracy as a POS tagger. The attempts started by Hajič [16], who built a training model to predict each word feature separately. Hajic used a direct dictionary as a source of possible morphological analyses (and hence tags) for an inflected word form. Statistical successful systems were presented by Ryan et.al. [28] for morphological disambiguation, Ibrahim et.al. [19], for mining parallel corpora to extract English-Arabic lemma pairs. The basic idea of theses researches is to extract all possible analysis, and corresponding features, for each single word using Buckwalter Arabic Morphological Analyzer (BAMA). Features are then fed to a classifier, which is trained on data from Arabic Tree Bank (ATB), to disambiguate the word category. Correct features including lemma form are then extracted from a BAMA lexicon. The approach yields excellent results for previously seen words. Stanford Arabic POS tagger is another supervised system. It depends on different trained models for many languages including Arabic. The accurate model for Arabic, was trained on whole ATB for maximum entropy [30].

## 3. Features of the Proposed approach

Although there are many approaches for stemming Arabic, no single approach is considered as a standard IR-oriented stemming algorithm. In the current research we aim at building a lemmatizer with minimum sufficient resources for IR purposes. Lemmatizer transforms inflected word form to its dictionary lemma look-up form. For nouns and adjectives, lemma form is the singular indefinite (masculine if possible) form, and for verbs, it is the perfective third person masculine singular form.

The basic idea is to collect more information about the word to be stemmed and its context to generate more accurate word features including its POS tags. The system exploits Arabic language knowledge in terms of roots, patterns, affixes, and a set of morpho-syntactic rules to generate accurate lemma form and its relevant morpho-syntactic features that support information retrieval purposes. Morpho-syntactic features are required also, to capture the important semantic senses of the language. Inflected languages such as Arabic, Finnish, Turkish, and Hungarian typically express meanings through morphological affixation. In highly Inflected languages plural, possessive relations, grammatical cases, and verb tenses, verb voice, and aspects which in English would be expressed with syntactic structures, are characteristically represented with morphological affixation [25]. Collected information about words also include word pattern. Arabic stem-patterns have interesting semantic features that give rise to senses of words. For example, syntactic patterns can recognize a given word as being the agent of an action, the instrument of that action, or the place at which the event occurs.

To implement our approach, the following features are considered:

- The proposed approach gets benefits of the power and generality of rule-based stemmers, and the accuracy of dictionary based approaches, which deals efficiently with cases of irregularity between singular and plural nouns, and proper nouns. Limited sized auxiliary dictionaries are used to enhance the performance of the lemmatizer.

- Relying only on lexicons may lead to much ambiguous word analysis possibilities. Therefore, Arabic context morpho-syntactic rules are used to expect the correct word category, and then verified. For example, the algorithm uses the word pattern and the category of its previous word for identifying the current word.

- In the current system, morphological and syntactic rules are adopted to reduce Arabic word into its abstract lemma form. Lemma is proved to be the smallest form that captures all semantic features of the word, and more suitable for IR systems.

- In spite of its importance in Arabic constructs, adjective is not classified as a POS main category of almost all Arabic POS taggers algorithms. Actually, traditional Arabic does not include adjective as one of its main parts-of-speech. An adjective in Arabic is actually a noun that happens to describe something. Many IR systems require aligning Arabic constructs with other languages constructs. Also, many IR systems extract candidate word category sequences that aid summarization and keyphrase extraction. In the proposed system, simple rules are used to re-categorize nouns as adjectives.

- The proposed system do not identify only word prefixes, suffixes and infixes, but finds out the corresponding morpho-syntactic attributes suitable for IR purposes at the lemma level.

The proposed algorithm has two main phases;

- POS tagging phase,
- Lemma generation phase.

## 4. POS tagging phase

There are many morphological analyzers for Arabic; some of them are available as an open source for research and evaluation, while the rest are proprietary commercial applications. Instead of "reinventing the wheel", we started our analysis phase implementation with the open source Khoja stemmer [21]. To achieve the proposed lemmatizer features, many modifications both in data and basic algorithm flow were necessary to add Arabic knowledge.

Khoja morphological analyzer is a root based stemmer, which removes possible infixes of a word, finds corresponding matched pattern, and extracts the word root without POS tags. The list of roots consists of 3800 trilateral and 900 quad literal roots. Also, Khoja system recognizes a list of 168 Arabic stop words.

In our implementation of the analysis phase, the algorithm relies on using different knowledge resources of Arabic language: prefixes, suffixes, patterns, and rules. Limited size auxiliary dictionaries are used to augment morphological and syntactic rules in recognizing words, and resolving their ambiguity. The dictionaries include only words that are expected to fail in tagging by rules. In most cases, the ambiguity is due to the absence of the short vowels in the electronic Arabic documents, or non templatic word stems. The basic algorithm outline is shown in figure (1).

```
For each word (WO) Do
  Begin word_block
    Search a word in proper noun dictionary
    If exists POS = N, with features set, exit word_block.
    Check the existence in closed set word dictionary
    If found, POS = article with features set, exit word_block;..
    For each affix -longest first- Do
      Begin affix_block
        If affix cannot be removed from W then exit affix_block;
        Remove affix to extract the (W) form
        Check if (W) matches a pattern (P) with root R
        If (P) exists
         Begin POS_block
           Apply POS identification rules;
             If rules failed POS =N;
           Apply syntactic rules to detect Adjectives
         End POS_block;
      End  affix_block;
    End loop;
  End word_block;
    End loop
```

Fig. 1 Outline of the Proposed Algorithm.

In the first stage, the algorithm starts the analysis by checking closed set Arabic words. The total list include 346 Arabic closed set words categorized into 16 groups (eg., prepositions, conjunctions, adverbs, numerals, etc.). Proper noun dictionary is also scanned at earlier stage of the analysis as shown in figure (1). The algorithm basic flow removes the longest suffix and the longest prefix in turn. After every elimination process, the algorithm checks a list of 61 patterns, if matches a pattern, the root is extracted and verified by checking the list of 3829 tri-roots. Up to this stage, the output is the suffix, prefix, word pattern, and root.

The purpose of the second stage is to tag POS of the word and to extract the corresponding features. The features for nouns and adjectives are definite case, count, and gender. Verbs extracted features are tense and voice. Finally, particles have 16 different subcategories. POS tagging and feature extraction are completed through many levels. The following subsections describe each level.

### 4.1 Identifying Nouns and Verbs

In Arabic language, some verbs or inflected nouns can have the same orthographic form due to absence of vowels. However, the algorithm exploits many techniques to disambiguate Arabic lexical categories. Words are identified at different levels through our algorithm as follows:-

a) The first level occurs after recognizing Arabic closed set words The existence of closed words such as ( بعد – فوق – أمام -إلى – ....) suggests that the next word is a noun. Also, The existence of closed words such as ( كلما–لم- لن- سوف– عندما .) suggests that the next word is a verb.

b) The second level is the syntax rules. For example, one rule states that if the previous word is a verb, the current word cannot be also a verb, since Arabic language does not permit two successive verbs to exist

c) Third level occurs during morphological stemming. In the proposed algorithm, affixes are categorized into three classes: affixes used by nouns only, affixes used by verbs only, and those that are used by either nouns or verbs. The existence of prefixes such as (لل ، ب ، ك، كال، ال), indicate that the word is a noun. Suffixes such as ( وه – ني – نهم –وا -) indicate that the word is verb. Word tag of the first two classes is well defined, while a word that has prefix and/ or suffix of the third class is still ambiguous.

d) The fourth level is the pattern-level, as illustrated in next subsection.

e) The remaining words after verbs identification are considered nouns.

### 4.2 Pattern level POS tagging

In our work, patterns play an important role in recognizing lexical word category. As shown in table (1). We classify Arabic patterns into three classes:

Table 1: Verb. Noun, and General patterns

| Pattern Class | Pattern | Form | Examples |
|---|---|---|---|
| Verb Patterns | انفعل<br>استفعل | en123<br>est123 | انتبه<br>استقام |
| Noun Patterns | مفعول<br>فعال<br>افتعال | m12ol<br>12a3<br>e1t2a3 | مكتوب<br>كتاب<br>اكتساب |
| General Patterns | فاعل<br>تفاعل<br>فعل | 1a23<br>t1a23<br>123 | ساعد V – شاعر N<br>تظاهر V – تضارب N<br>كتب V - كتب N |

1- Verb Patterns: which are used for verbs only.
2- Noun Patterns: which are used for nouns only.
3- General Patterns: which may be used for verbs or nouns according to different vocalization and not-written diacritics

If the word pattern belongs to first or second classes, it is a straight forward task to tag it as a noun or a verb respectively. For example, the word "ضوابط" corresponds to the pattern "فواعل", and hence the POS feature of the word is set to noun.

Words that have patterns belonging to third class cannot be tagged unless having a dictionary. In our implementation, instead of storing all Arabic verb forms (abstract and augmented), we store only most common verbs that belong to third class patterns. The current dictionary includes 943 common verbs, belonging to third class.

### 4.2 Identifying Adjectives

Traditionally, Arabic does not include adjective as one of its main POS. An adjective in Arabic is actually a noun that happens to describe something. Adjectives take almost all the morphological forms of nouns. Adjectives, for example, can be definite, and are inflected for case, number and gender. In this stage of analysis noun words are checked to see if it is actually an adjective. The algorithm uses a shallow level sentence structure for this stage. In the proposed lemmatizer, the Noun category of a word is changed to an adjective if the following conditions are met

1- The current word does not contain any prefix.
2- Its previous word is also a noun (or adjective) and has the same count and gender.
3- Both of current and previous words are definite or both of them are indefinite.

In our implementation, the feature definite is different from [DET] used by Ryan et.al. [28], which checks only the existence of definite determinant. For example, Arabic nouns with attached possessive pronouns are definite in spite of non-existence of definite determinant**.**

## 5. Lemma Generation phase

The main contribution of this work is the development of lemma generator that extracts the lemma form of an Arabic word. On a word form conflation scale, lemma representation lies slightly above the (minimum) stem level, and below the (maximum) root level. The purpose of the second phase of the lemmatizer algorithm is to generate the abstract lemma form of a word. The next two subsections describe the procedures used to generate verb's and noun's lemmas.

### 5.1 Generating Verb's lemma

Verb lemma is the perfective, 3rd person, singular verb form In most cases, lemma is the same as the root form of a verb. For example, the word (يكتب) has same root and lemma form (كتب). In other cases, removing prefixes and suffixes is not enough to generate the lemma form. In our implementation, rules are applied at the pattern level to deal with such cases. Table (2) shows examples of the analysis. The table shows that lemma form may be different from root form of a verb, and it may be required to remove or substitute prefix of the word stem.

Table 2: Examples of differences between pattern and lemma forms for verbs

| Initial Word | After removing | Root | Pattern | Pattern form | Lemma form |
|---|---|---|---|---|---|
| تنازل | تنازل | نزل | تفاعل | t1a23 | تنازل |
| يستخرجون | يستخرج | خرج | يستفعل | yst123 | استخرج |
| نحتاجهم | نحتاج | حوج | نفتعل | n1t23 | احتاج |
| تندرج | تندرج | درج | تنفعل | tn123 | اندرج |

### 5.1 Generating Noun's lemma

Lemma form of noun (or adjective) is the singular indefinite form. In Arabic, there are two types of noun and adjective plural forms: regular plurals, and broken (irregular) plurals. Regular plural can be masculine plural and feminine plural.

**Regular Plural:** The lemma form of the masculine plural is generated simply by removing prefixes "ون" or "ين" from a noun form. Lemma singular form of feminine plural nouns has two cases; feminine or masculine single form. Removing the suffix "ات" is enough for masculine single form case. Feminine singular form requires, adding "ة" to indicate its feminine nature. In our implementation, a dictionary is used to store feminine single noun forms. The algorithm checks the noun word in the dictionary of the feminine words, if exists a character 'ة' is added.

Also, feminine have suffixes composed of 'ت' and attached pronoun (e.g., words وظيفتك، معالجتها), requires suffix substitution when generating the lemma form. The rule is to replace the end character 'ت' with character 'ة', after removing the attached pronoun suffixes (lemma form will be معالجة and وظيفة).

**Broken Plural nouns:** Another problem with nouns and adjectives lemma generation is the issue of broken plurals. The term was chosen to indicate that the base form of the nouns is broken either by removing one or more letters, adding one or more letters, changing vocalization or a combination of these. There are about 27 pattern forms for the broken plural [31], and in many cases, to generate the singular form, there are a lot of probabilities for the singular form pattern. For example if the broken plural pattern is (فعلاء) the singular form pattern may be one of the two patterns (فاعل ، فعيل), Also, the broken plural words (جهلاء ، كرماء) each of them has a pattern (فعلاء), and have different single forms (جاهل، كريم ), which corresponds to the patterns (فاعل، فعيل) respectively, and there is no rule to determine exactly which of them is correct.

Therefore, it is very difficult to rely only on morphological rules without a dictionary to guess the lemma form of these broken plural ambiguous cases. In the proposed work, a dictionary is used to store only ambiguous cases, i.e., that have a lot of probabilities for the singular form.

## 6. Performance measurements

In order to evaluate the performance of the proposed system, two experiments were carried out on two different datasets. In the first experiment, a dataset is used to tune and evaluate the overall performance of the algorithm. In the second experiment, another dataset is used to compare the accuracy of the proposed approach and an up to date Stanford POS for unseen before documents. The following sub sections describe the details of each experiment

### 6.1 Experiment 1: Maximizing Performance

In the first experiment, a dataset was used to maximize the performance, and to measure the algorithm upper and lower accuracy bounds. The dataset contains 50 documents with 32,500 manually annotated words. The data are collected from different online resources and domains with focus on technical documents. It includes journal articles, technical reports, papers, and sections from text books. The first dataset is manually tagged to main POS (Nouns, Adjectives, Verbs, Articles, Proper Nouns, and Unknown). Unknown word category includes misspelling and non-

Arabic words. Table (3), shows the dataset POS tags as a percentage of the total words of the documents.

The purpose of the first experiment is to measure how much accuracy can be achieved with the proposed lemmatizer for document. Therefore, through the first experiment, some rules are rewritten, and words are added to maximize the performance. It is found that the proposed lemmatizer can achieve an accuracy of 94.8% of known documents. The efficiency is determined by the number of exact POS matches between manually tagged dataset words, and the proposed approach tagged words, divided by the total dataset words.

Table 3: POS tags for the first dataset

| Word POS | % of total words |
|---|---|
| Nouns | 50.0% |
| Verbs | 8.4% |
| Adjectives | 9.3% |
| Proper Nouns | 3.7% |
| Closed System | 25.9% |
| Unknown | 2.7% |

Through the experiment, different algorithm procedures are monitored to investigate their effect on the overall efficiency. The lower bound corresponds to running the algorithm with minimum resources. Through the experiment, it is found that using very simple procedures can lead to a minimum efficiency limit of 70% in case of ambiguous words. Storing only a look up dictionary of Arabic closed word set, simple definite prefixes, suffixes, and adjective rules, allow the algorithm to detect most of the nouns and adjectives successfully. The details of participation of each procedure are given below:

1) Looking up closed system Arabic words (particles), recognizes 25.9% of the total document words.
2) Using simple techniques for recognizing nouns leads to recognition of 75.4% of total nouns in the dataset, which corresponds to a 37.7% of the total document words. The details of these techniques are:
   a. 28.2% of the document words are initially categorized as nouns by detecting the presence of definite prefixes "ال". Actually, this includes both definite nouns and definite adjectives:
      i. Applying simple syntactic rules of adjective - only definite successive noun rule - allows tagging of 6.8% of total document words as adjectives. This corresponds to 73.1% of total adjectives in our dataset,
      ii. The remaining 21.4% of definite document words are actual nouns.
   b. 9.1% of the document words are succeeded to be categorized as nouns by detecting the presence of definite prefixes "لل".
   c. 2.5% of the document words are categorized as nouns by detecting the presence of prefixes "ك" and "ب".
   d. An additional 4.6% of the document words are categorized successfully as nouns by applying previous word category rules.

Table 4: Sample of results of the first experiment

| Stanford Stemmer | Proposed Lemmatizer | | | | English | Arabic word |
|---|---|---|---|---|---|---|
| POS | R | P | L | POS | | |
| VBP | عمد | تفتعل | اعتمد | VV | It (female) depends | تعتمد |
| NOUN | | | معظم | particle | most | معظم |
| NN | بلد | فعلان | بلد | NNS | countries | بلدان |
| DTNN | علم | فاعل | عالم | DTNN | the world | العالم |
| DTNN | | | الآن | RB | now | الآن |
| IN | | | على | IN | on | على |
| NN | خدم | استفعال | استخدام | NN | use | استخدام |
| DTNN | نظم | افعل | نظام | DTNNS | the systems | الأنظمة |
| DTJJ | بني | مفعلة | مبني | DTJJ | based | المبنية |
| IN | | | على | IN | on | على |
| DTNN | حسب | فاعل | حاسب | DTNN | the computer | الحاسب |
| DTJJ | | | آلي | DTJJ | the automatic | الآلي |
| IN | | | في | IN | in | في |
| NN | نشأ | افعال | انشاء | NN | building | إنشاء |
| NN | شغل | تفعيل | تشغيل | NN + | and operating | وتشغيل |
| NN | | | صيانة | NN+ | and maintenance | وصيانة |
| NN | شرع | مفاعيل | مشروع | NNS | projects | مشاريع |
| DTNN | بني | فعلة | بنية | DTNN | the infra | البنية |
| DTJJ | سوس | افعل | اساس | DTJJ | the basic | الأساسية |
| DTJJ | خوص | فعلة | خاصة | DTJJ | the dedicated | الخاصة |
| NN | | | بها | particle | for it | بها |
| IN | | | في | IN | in | في |
| NN | خلف | مفتعل | مختلف | NN | different | مختلف |
| DTNNS | قطع | فعال | قطاع | DTNNS | the sectors | القطاعات |
| NN | مثل | | مثل | NN | like | مثل |
| NNS | قطع | فعال | قطاع | JJ | sectors | قطاعات |
| DTNN | صنع | فعالة | صناعة | DTNN | the industry | الصناعة |
| NN | زرع | فعالة | زراعة | DTNN + | and the agriculture | والزراعة |
| NN | | | صحة | DTNN + | and the health | والصحة |
| NNP | علم | تفعيل | تعليم | DTNN + | and the education | والتعليم |
| NNP | تجر | فعالة | تجارة | DFNN + | and the commerce | والتجارة |
| NNP | بنك | فعول | بنك | DTNNS + | and the banks | والبنوك |
| NNS | خدم | | خدمة | DTNNS + | and the services | والخدمات |
| JJ . | | | غير | particle | And others | وغيرها، |
| PUNC | | | | PUNC | | . |

Hence, in this minimum resource experiment, approximately 70% of total Arabic document words can be tagged without use of any roots or patterns as resources. This is the minimum boundary of the algorithm, which is equivalent to light stemmers operation. However, this minimum accuracy level is not accepted by many IR systems, since until now, no verbs are detected.

6.1 Experiment 2: Comparing Results

In the second experiment, we compare our root-based Arabic Lemmatizer output to Stanford POS Arabic tagger [30], with trained model (Arabic-accurate model dated 2011-09-14). The Stanford model was trained on the entire ATB p1-3. A second dataset is used through this experiment, which is a non-technical Arabic document collected for building Contemporary Arabic Corpus [4], The document contains 4020 Arabic words in education domain.

```
KAN/كان النظام/DTNN التعليمي/DTJJ في/IN العراق/DTNNP من/IN أكثر/NN
النظم/DTNNS تقدمًا/particle في/IN / العالم/DTNN العربي/DTJJ قبل/ADV
عام/NN 1990./NUM بيد/NN أن/particle هذا/DEMO النظام/DTNN تدهور/VV
تدهورًا/VV كبيرًا/NN نتيجة/NN الحروب/DTNNS التي/CONJ تورط/VV
فيها/particle النظام/DTNN السابق/DTJJ وما/NN أعقبها/JJ من/IN فرض/NN
عقوبات/NNS دولية/JJ على/IN البلاد/DTNNS مما/particle أدخلها/NN في/IN
دائرة/NN الإهمال/DTNN والانعزال/DTNN + وأورث/NN مشكلات/NNS ضخمة/JJ
ما/particle زالت/VV البلاد/DTNNS تعانيها/NN في/IN / الوقت/DTNN
الحالي./particle وقد/حرف تفاقمت/VV الأوضاع/DTNNS نتيجة/NN أعمال/NNS
التدمير/DTNN والنهب/NN والتعطيل/DTNN + لمؤسسات/NNS
الدولة،/DTNN والتي/CONJ وقعت/VV منذ/IN شهر/ADV مارس/NN ٢٠٠٣م/NUM
في/IN أعقاب/NNS سقوط/JJ العاصمة/DTNN بغداد/NNP وانهيار/NN
النظام/DTNN السياسي/DTJJ ودخول/NN القوات/DTNNS الأمريكية/unkown
والبريطانية/unkown للبلاد.+ DTNNS ويأمل/VV المجتمع/DTNN الدولي/DTJJ
بعد/ADV أن/particle يستتب/unkown الأمر/DTNN في/IN / العراق/DTNNP
ويؤول/VV الحكم/DTNN ومقاليد/NNS السلطة/DTNN لحكومة/NN + وطنية/JJ
عراقية/+DTNNS منتخبة/NN أن/particle يتحرك/VV العراق/DTNNP بسرعة/NN
+ لإعادة/NN بناء/NN النظام/DTNN التعليمي/DTJJ وتأهيله/NN + وتجديده./+NN
```

Fig. 2 Output of proposed lemmatzer, 15 mistakes were detected out of 110 words (non-technical document). Mistakes are shown bolded.

```
كان/VBD النظام/DTNN التعليمي/DTJJ في/IN العراق/DTNNP من/IN أكثر/NN
النظم/DTNN تقدمًا/NN في/IN العالم/DTNN العربي/DTJJ قبل/NN عام/NN
NN/بيد NN/أن DT/هذا NN/النظام DTNN/تدهور NN/تدهورًا
كبيرًا/NN نتيجة/NN الحروب/DTNN التي/WP تورط/VBP فيها/NN النظام/DTNN
السابق/DTJJ وما/NN أعقبها/NN من/IN فرض/NN عقوبات/NNS دولية/JJ على/IN
البلاد/DTNN مما/NN أدخلها/NN في/IN دائرة/NN الإهمال/DTNN والانعزال/NN
وأورث/NN مشكلات/NNS ضخمة/JJ ما/RP زالت/VBD البلاد/DTNN تعانيها/NN
في/IN الوقت/DTNN الحالي/DTJJ PUNC/. وقد/NN تفاقمت/VBD الأوضاع/DTNN
نتيجة/NN أعمال/NN التدمير/DTNN والنهب/NN والتعطيل/NN لمؤسسات/NNS
الدولة/DTNN ،/NNS والتي/VBD وقعت/VBD منذ/IN شهر/NN مارس/NN
2003م/CD في/IN أعقاب/NN سقوط/NN العاصمة/DTNN بغداد/NNP وانهيار/NNP
النظام/DTNN السياسي/DTJJ ودخول/NN القوات/DTNNS الأمريكية/DTJJ
والبريطانية/NN للبلاد/PUNC . NNP/ويأمل المجتمع/DTNN الدولي/DTJJ
بعد/NN أن/NN يستتب/VBP الأمر/DTNN في/IN العراق/DTNNP ويؤول/NN
الحكم/DTNN ومقاليد/NN السلطة/DTNN لحكومة/NN وطنية/JJ عراقية/JJ منتخبة/JJ
أن/VBD يتحرك/VBP العراق/DTNNP بسرعة/NN لإعادة/NN بناء/NN
النظام/DTNN التعليمي/DTJJ وتأهيله/NNP وتجديده/PUNC . NNP/
```

Fig. 3 Output of Stanford Arabic stemmer, 28 mistakes were detected out of 110 words (non-technical document). Mistakes are shown bolded.

In our experiment on the unseen before dataset, the average accuracy of the proposed algorithm is 89.15% as a POS tagger, while it is 76.7% for Stanford Arabic accurate model. Figure (2), and figure (3), show the output of both algorithms for the same sample text. In our algorithm, most of the errors encountered result from undiscovered verbs, proper nouns, and confusing nouns with adjectives, Stanford Arabic stemmers errors are mainly due to improper detection of Arabic broken plurals, and limited coverage of basic Arabic words.

In an error analysis of Stanford POS tagger [26], it is noted that 4.5% of errors are due to unknown word, where the tagger has to rely only on context features, and contexts are often ambiguous. Stanford tagger for Arabic complaints real problems with categorizing broken plural nouns, definite nouns detection, and noun-verb confusion.

## 7. Conclusions

In this research we have presented an accurate algorithm for extracting lemma form of Arabic words and their morpho-syntactic features. The presented algorithm proves that accurate results for POS tagging, can be achieved when using inherent features and rules of Semitic languages like Arabic. It is shown that ambiguity can be resolved using metadata about patterns, roots, and infixes' indications. Analysis is aided with auxiliary dictionaries and syntax rules to produce a lemmatizer which outperforms exciting up to date Arabic learning algorithms. .

**Tarek El-Shishtawy** is a Professor assistant at Faculty of Engineering, Benha University, Egypt. He participated in many Arabic computational Linguistic projects. Large Scale Arabic annotated Corpus, 1995, was one of important projects for Egyptian Computer Society, and Academy of Scientific Research and Technology, He has many publications in Arabic Corpus, machine translation, Text, and data Mining.

**Fatma El-Ghannam** is a researcher assistance at Electronics Research Institute – Cairo, Egypt. She has great research interests in Arabic language generation and analysis. Currently, she's preparing for a Ph.D. degree in NLP.